\documentclass[sigconf]{acmart}

\usepackage{subfig}

\usepackage{balance}

\AtBeginDocument{%
  \providecommand\BibTeX{{%
    \normalfont B\kern-0.5em{\scshape i\kern-0.25em b}\kern-0.8em\TeX}}}

\begin{document}
\fancyhead{} 

\copyrightyear{2020}
\acmYear{2020}
\setcopyright{acmcopyright}\acmConference[WebSci '20]{12th ACM Conference on Web Science}{July 6--10, 2020}{Southampton, United Kingdom}
\acmBooktitle{12th ACM Conference on Web Science (WebSci '20), July 6--10, 2020, Southampton, United Kingdom}
\acmPrice{15.00}
\acmDOI{10.1145/3394231.3397890}
\acmISBN{978-1-4503-7989-2/20/07}

\title{DeepHate: Hate Speech Detection via \\Multi-Faceted Text Representations}


\author{Rui Cao}
\affiliation{%
  \institution{University of Electronic Science and Technology of China, China}
 }
\email{caorui0503@gmail.com}

\author{Roy Ka-Wei Lee}
\affiliation{%
  \institution{University of Saskatchewan, Canada}
 }
\email{roylee@cs.usask.ca}

\author{Tuan-Anh Hoang}
\affiliation{%
  \institution{Leibniz University of Hanover, Germany}
 }
\email{hoang@l3s.de}


\begin{abstract}
Online hate speech is an important issue that breaks the cohesiveness of online social communities and even raises public safety concerns in our societies. Motivated by this rising issue, researchers have developed many traditional machine learning and deep learning methods to detect hate speech in online social platforms automatically. However, most of these methods have only considered single type textual feature, e.g., term frequency, or using word embeddings. Such approaches neglect the other rich textual information that could be utilized to improve hate speech detection. In this paper, we propose \textsf{DeepHate}, a novel deep learning model that combines multi-faceted text representations such as word embeddings, sentiments, and topical information, to detect hate speech in online social platforms. We conduct extensive experiments and evaluate \textsf{DeepHate} on three large publicly available real-world datasets. Our experiment results show that \textsf{DeepHate} outperforms the state-of-the-art baselines on the hate speech detection task. We also perform case studies to provide insights into the salient features that best aid in detecting hate speech in online social platforms.
\end{abstract}




\maketitle

\section{Introduction}
\label{sec:introduction}
\textbf{Motivation.} The proliferation of social media has enabled users to share and spread ideas at a prodigious rate. While the information exchanges in social media platforms may improve an individual's sense of connectedness with real and virtual communities, these platforms are increasingly exploited for the propagation of toxic content such as hate speeches \cite{schmidt2017survey,fortuna2018survey}, which are defined by the Cambridge dictionary as ``\textit{public speech that expresses hate or encourages violence towards a person or group based on something such as race, religion, sex, or sexual orientation }'' \cite{cambridge}. The spread of hate speech in social media has not only sowed discord among individuals or communities online but also resulted in violent hate crimes \cite{Williams19,relia2019race,mathew2019spread}. Therefore, it is a pressing issue to detect and curb hate speech in online social media.

Major social media platforms such as Facebook and Twitter have made great efforts to combat the spread of hate speech in their platforms \cite{Times19, bloomberg19}. For instance, the platforms have provided clear policies on hateful conducts \cite{facebookrules,twitterules}, implemented mechanisms for users to report hate speech, and employed content moderators to detect hate speeches actively. However, such approaches are labor-intensive, time-consuming, and thus not scalable or sustainable in the long run \cite{waseem2016hateful,gamback2017using}.  

The gravity of the issues and limitations of manual approaches has motivated the search for automatic hate speech detection methods. In recent years, researchers from data mining, information retrieval, and Natural Language Processing (NLP) fields have proposed several such methods  \cite{fortuna2018survey,schmidt2017survey}. These methods can be broadly grouped into two categories: (i) methods that adopted classic machine-learning strategies \cite{chen2012detecting,waseem2016hateful,waseem2016you,nobata2016abusive,chatzakou2017mean, DavidsonWMW17}, and more recently, (ii) deep learning-based methods \cite{djuric2015hate,mehdad2016characters,gamback2017using,badjatiya2017deep,ParkF17,grondahl2018all,zhang2018detecting,arango2019hate,FountaCKBVL19}. 

For traditional machine learning-based methods, textual features such as bag-of-words are commonly extracted from social media posts to train a classifier to detect hate speech in the posts. For deep learning-based methods, words in the posts are often represented by some word embedding vectors and fed as inputs to train a neural network for hate speech detection. Although the existing methods, especially the deep learning ones, have shown promising results in automatic hate speech detection in social media, there are limitations in these models. Firstly, most of the existing methods have only considered single type textual features, neglecting the other rich textual information that could be utilized to improve hate speech detection. Secondly, the current deep learning methods offered limited explainability on why a particular post should be flagged as hate speech. 

\textbf{Research Objectives.} In this paper, we address the limitations in existing methods and propose \textsf{DeepHate}\footnote{Code: https://gitlab.com/bottle\_shop/safe/deephate}, a novel deep learning architecture that effectively combines multi-faceted textual representations for automatic hate speech detection in social media. At a high level, \textsf{DeepHate} utilizes three types of textual representations: semantic, sentiment, and topical information. The semantic representations of social media posts are extracted from several word embeddings pre-trained on a much larger text corpus \cite{mikolov2018advances, PenningtonSM14, wieting2015paraphrase}. For sentiment representation, we proposed a two-step approach that utilizes an existing sentiment analysis tool \cite{HuttoG14} and neural network to train a new word embedding that captures the sentiment information in a social media post. Latent Dirichlet Allocation (LDA) \cite{blei2003latent} is used to extract topical representations of posts. The three types of textual representations are subsequently used as input to train our \textsf{DeepHate} model for hate speech classification. The underlying intuition is that the multi-faceted textual representations enrich the representation of social media posts, and this enables the deep learning classifier to perform better hate speech detection. To gain a better understanding of how \textsf{DeepHate} decides which posts to classify as hate speech, we also conduct empirical studies to analyze the salient features aided in hate speech detection.

\textbf{Contributions.} Our main contributions in this work consist of the following.
\begin{itemize}
    \item We propose a novel deep learning model called \textsf{DeepHate}, which utilizes multi-faceted textual representations for automatic hate speech detection in social media.
    \item We conduct extensive experiments on three real-world and publicly available datasets. Our experiment results show that \textsf{DeepHate} consistently outperforms state-of-the-art methods in the hate speech detection task. 
    \item We conduct empirical analyses on the \textsf{DeepHate} model and provide insights into the salient features that helped in detecting hate speech in social media. The salient feature analysis also improves the explainability of our proposed model. 
\end{itemize}

\section{Related Work}
\label{sec:related}
Automatic detection of hate speech has received considerable attention from the data mining, information retrieval, and natural language processing (NLP) research communities. Interest in this field has increased with the proliferation of social media and social platforms. In this section, we focus on reviewing the existing works for detecting hate speech in social media text content. Mostly, we will be focusing on hate speech detection on Twitter short messages (i.e., tweets). These works can be broadly categorized into two  approaches: (i) works that adopt classic machine-learning strategies, and, more recently, (ii) those that deep learning methods. Readers are encouraged to refer to recent the surveys on the topic \cite{fortuna2018survey,schmidt2017survey} for great detail of methods for both Twitter and other social media.

Traditional machine learning methods have been applied to detect hate speech in social media \cite{chen2012detecting,waseem2016hateful,waseem2016you,nobata2016abusive,chatzakou2017mean, DavidsonWMW17}. Typically, these methods include an initial feature extraction phase, where features are extracted from the raw textual content. The most commonly extracted features include Term Frequency Inverse Document Frequency (TF-IDF) scores, Bag-of-Words vectors, and other linguistic attributes. Xiang et al. \cite{xiang2012detecting} had also explored the latent semantic features extracted from tweets for hate speech detection by mining topics of the tweets. Beyond the textual content, some studies have also utilized other meta-information from the users' profiles and network structures (i.e., followers, mentioned, etc.) \cite{chatzakou2017mean,papegnies2017graph,singh2017toward}. The extracted features are subsequently used as input for classifiers such as Logistic Regression, SVM, Random Forest, etc., to predict if the given tweet contains hate speeches. In this paper, we focus on utilizing only the textual content to perform hate speech detection as other meta-information is scarce and often incomplete. Moreover, by not using these meta-information, we would avoid to bias our hate detectors by any personal information of the users.      

Deep learning methods have achieved notable performance in many classification tasks. Unlike traditional machine learning methods, deep learning methods are able to automatically learn latent representations of the input data to perform classification \cite{goodfellow2016deep}. Such deep learning approaches have also been applied to several natural language processing tasks, including text classification \cite{goldberg2016primer,yang2016hierarchical}. The increasing popularity of deep learning approaches also sees a number of recent studies adopting these methods to detect hate speech in social media \cite{djuric2015hate,mehdad2016characters,gamback2017using,badjatiya2017deep,ParkF17,grondahl2018all,zhang2018detecting,arango2019hate,FountaCKBVL19}. 

Mehdad et al. \cite{mehdad2016characters} experimented applying Recurrent Neural Network (RNN) model different input types such as word, unigram, and bigram character embeddings for hate speech detection. Gamb{\"a}ck et al. \cite{gamback2017using} conducted similar studies using Convolutional Neural Network (CNN). Badjatiya et al. \cite{badjatiya2017deep} proposed an ensemble approach that combines Long-Short Term Memory (LSTM) model and Gradient-Boost Decision Tree to detect hate speech on Twitter. Park and Fung \cite{ParkF17} introduced the \textit{HybridCNN} method that trains a CNN over both word and unigram character embeddings for hate speech detection. Zhang et al. \cite{zhang2018detecting} proposed a new neural network architecture that combines CNN with Gate Recurrent Unit (GRU) to classify hate speech using word embeddings as input. Founta \cite{FountaCKBVL19} trains a combined RNN and Multilayer perceptron (MLP) on textual content and meta-information to perform hate speech detection. Most of these studies have applied and experimented with their proposed methods on datasets proposed in \cite{waseem2016hateful,DavidsonWMW17}. Unlike the existing deep learning methods, which mainly utilized word or character embeddings as input, we proposed a novel deep learning architecture to combine multi-faceted text representations for the classification of hate speech. We will evaluate and benchmark our proposed model against these state of the art in Section \ref{sec:experiment}.

\section{Proposed Model}
\label{sec:model}
\begin{figure}[ht!]
		\includegraphics[width=0.45\textwidth]{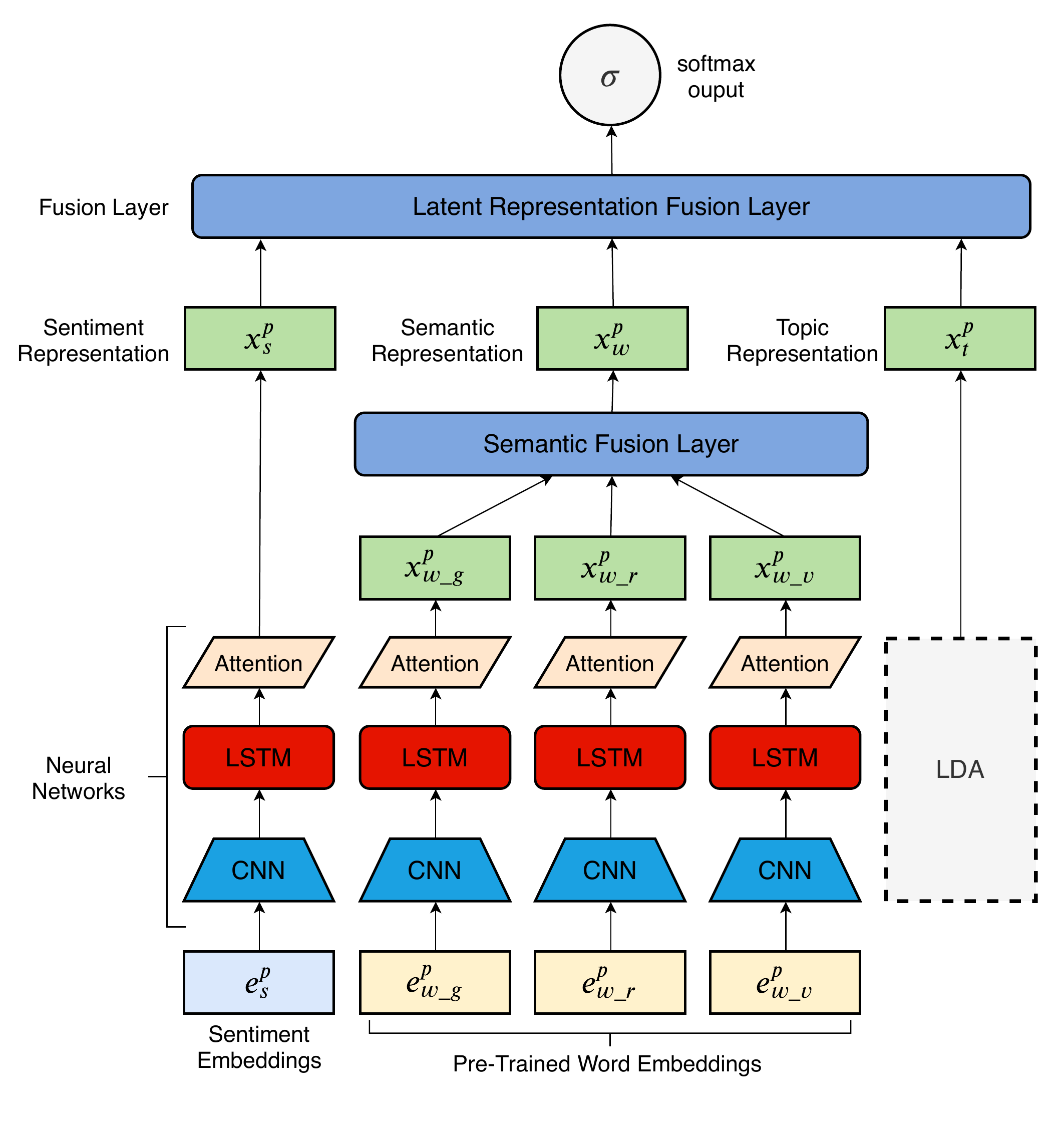}
	\caption{Overall architecture of \textsf{DeepHate} model}
	\label{fig:full-model}
\end{figure}

In this section, we elaborate on our proposed \textsf{DeepHate} model. The intuition of our proposed model is to learn the latent representations of multi-faceted text information, and effectively combine them to improve hate speech detection. Figure \ref{fig:full-model} illustrates the overall architecture of the \textsf{DeepHate} model. The proposed model first utilizes different types of feature embeddings to represent a post $p$. The feature embeddings are subsequently fed into neural network models to learn three types of textual information latent representations, namely, semantic, sentiment, and topic. The latent representations are then combined via a feed-forward network. Finally, a softmax layer takes the combined representation as input to predict probability distribution over all possible classes. The details of the individual components in \textsf{DeepHate} model are described in the subsequent subsections.


\subsection{Semantic Representation}
To learn the latent semantic representation of a post $p$, we first represent the post as a sequence word, i.e., $W^{p} = (w^{p}_1,w^{p}_2,\cdots,w^{p}_L)$, where $L$ refers to the length of the post. Each word in the sequence is then represented by pre-trained word embeddings. In order to obtain expressive word-level representations, we use three popular pre-trained word embeddings, namely: GloVe~\cite{PenningtonSM14}, word2vec trained on wiki news~\cite{mikolov2018advances}, and Paragram \cite{wieting2015paraphrase}. Consequently, we denote the pre-trained word embeddings of post $p$ as follows:

\begin{itemize}
    \item \textit{Glove}: $e^{p}_{w\_g} = (e^{p}_{w\_g_1},e^{p}_{w\_g_2},\cdots,e^{p}_{w\_g_L})$
    \item \textit{Word2Vec-Wiki}: $e^{p}_{w\_v} = (e^{p}_{w\_v_1},e^{p}_{w\_v_2},\cdots,e^{p}_{w\_v_L})$
    \item \textit{Paragram}: $e^{p}_{w\_r} = (e^{p}_{w\_r_1},e^{p}_{w\_r_2},\cdots,e^{p}_{w\_r_L})$
\end{itemize}

Note that all pre-trained word embeddings are trained as 300-dimensional vectors. Each pre-trained word embedding of a post is used as input into a C-LSTM-Att encoder to learn a latent representation of the post. For instance, we learn the latent post representations $x^p_{w\_g}$ from $e^p_{w\_g}$, $x^p_{w\_v}$ from $e^p_{w\_v}$, and $x^p_{w\_r}$ from $e^p_{w\_r}$. The detail implementation of the C-LSTM-Att encoder will be presented in section \ref{sec:cnn_lstm}. 

Finally, we generate the post's latent semantic representation, $x^p_w$, by combining the three latent post representations learned from various types of pre-trained word embeddings. We want to fully exploit the advantage of each kind of representation. As a consequence, we assign an attention weight, which is a single value, to each vector of representation. To make it easier for training, these attention weights $a_{g}, a_{v}, a_{r}$ are initialized by standard uniform distribution.
Element-wise summation are performed on the three attended latent post representation vectors.
\begin{equation}
    x^{p}_{w} = a_{g}x^p_{w\_g} + a_{v}x^p_{w\_v} + a_{r}x^p_{w\_r}
\end{equation}

$x^p_w$ will be subsequently combined with other textual information latent representations for hate speech detection in the fusion process described in Section \ref{sec:fusion}. 


\subsection{Sentiment Representation}
\label{sec:sentiment}
Tweets often contain abundant sentiment information \cite{giachanou2016like}. Each tweet encodes the attitude and emotion of the writer, and such information may be helpful in hate speech detection. From this intuition, we aim to incorporate sentiment information in the \textsf{DeepHate} model. Inspired by the method proposed in \cite{tang2014learning}, our goal here is to train a word embedding that contains sentiment information. Then, similar to the training of latent semantic representation, this trained sentiment-specific word embedding, denoted by $e^{p}_{s}$, will be used to represent the words in a post, which is subsequently used as input into the C-LSTM-Att encoder to learn the latent sentiment representation of the post.     

However, the technique proposed in \cite{tang2014learning} is a supervised method that requires labels for sentiment classification, and the existing hate speech datasets lack such sentiment labels. To overcome this limitation, we first explore a sentiment analysis tool, VADER~\cite{HuttoG14}, to label the sentiment for the tweets in the hate speech datasets. VADER is a lexicon and rule-based sentiment analysis tool that is specifically tuned to extract sentiments expressed in social media. Given a tweet, the tool generates scores over three polarities: negative, positive, and neutral. In our implementation, we assume the sentiment label for a tweet to be the polarity with the highest score. With the generated sentiment labels for the hate speech datasets, we train a sentiment-specific word embedding \cite{tang2014learning} by performing a sentiment classification task to predict the sentiment labels in hate speech datasets. Similar to the pre-trained word embeddings, the sentiment-specific word embedding (a.k.a sentiment embedding) is trained as 300-dimensional vectors. 

Finally, we represent the word sequence of a post using the trained sentiment embedding, $e^{p}_{s} = (e^{p}_{s_1},e^{p}_{s_2},\cdots,e^{p}_{s_L})$, and use it as input into a C-LSTM-Att encoder to learn a latent sentiment representation of the post, $x^{p}_{s}$. Note that we also freeze the parameters in the sentiment embedding layers such that the sentiment embedding is not updated; this ensures the sentiment information is preserved and not updated by the back-propagation from our hate speech detection task.

\subsection{Topic Representation}
We employ the probabilistic topic modeling approach \cite{blei2012probabilistic} to derive the topic representation of the posts. Generally, this approach allows to represent each text document by a multinomial distribution over topics where each topic is mathematically defined as a multinomial distribution over words. Often, in mining topics from Twitter posts, the documents are formed by aggregating the posts by hashtags, published time, or authors \cite{hong2010empirical,zhao2011comparing}. However, since our datasets are highly sparse in those dimensions (e.g., there are not many common hashtags among posts, posts are published in a long span of time, and each user has very few posts), we consider each single post is a  document and make use of the original Latent Dirichlet Allocation model \cite{blei2003latent} (LDA) for computing the posts' topic distribution. Moreover, inspired by previous works that each post should focus on one or a few topics, we exploit sparsity regularization in the learning of the LDA model \cite{balasubramanyan2013regularization}. Thus, posts' topic representations, $x^{p}_{t}$, are sparse vectors (which are highly skewed multinomial distribution over topics).

Specifically, we first remove all stopwords and non-English words. Next, we iteratively filter away infrequent words and too short posts such that: each word must appear in at least five remaining posts while each post contains at least three remaining words. These minimum thresholds are to ensure that for each post and each word, we have enough observations to learn the topics accurately. To determine the appropriate number of topics in each dataset, we run the LDA model (with sparsity regularization) with the number of topics varied from 5 to 20 and measure the likelihood of the learned model in each case. The number of topics in each dataset is then set by considering both computational complexity and improvement in the likelihood. Lastly, for too short posts that are previously filtered away, we assume that their topic distributions are uniform. There is just a small proportion (often less than 10\%) of such posts in each dataset.

\subsection{C-LSTM-Att Encoder}
\label{sec:cnn_lstm}
Zhou et al. \cite{zhou2015c} introduced the C-LSTM model, which stacked a convolutional neural network (CNN) and long-short term memory (LSTM) \cite{hochreiter1997long}, for text classification. Motivated by C-LSTM's good performance in the text classification task, we modify C-LSTM by adding an attention mechanism \cite{yang2016hierarchical} to select informative words that contribute more towards detecting hate speech in social media posts. Also, our proposed network can be adaptive to convolution operation with multiple filter sizes. We name the modified model \textit{C-LSTM-Att}. The C-LSTM-Att model is used to encode the post's pre-trained word embeddings, i.e., $e^p_{w\_g}$, $e^p_{w\_v}$, $e^p_{w\_r}$, and sentiment embedding, $e^p_{s}$, into its latent semantic and sentiment representations. As the process of encoding the four feature embeddings is similar, we use $e^p$ as a generic word embedding and $x^p$ as a generic latent semantics or sentiment representation of a post for illustration in our subsequent discussion in this section.

\textbf{CNN Component.} This component is a slight variant of the traditional convolutional networks \cite{goodfellow2016deep}. Let $e^p_i \in \mathbb{R}^d$ be the $d$-dimensional word vectors for the $i$-th word in a post $p$. Let $e^p \in \mathbb{R}^{L \times d}$ denote the input post where $L$ is the length of the post $p$. Let $k$ be the length of the filter, and the vector $\mathbf{m} \in \mathbb{R}^{k \times d}$ is a filter for the convolution operation. For each position $j$ in the post, we have a window vector $w_j$ with $k$ consecutive word vectors, denoted as:   

\begin{equation}
    w_j = [e^p_j, e^p_{j+1},\cdots,e^p_{j+k-1}]
\end{equation}

Where the commas represent row vector concatenation. A feature map $\mathbf{c} \in \mathbb{R}^{L-k+1}$ is generated with as the filter $\mathbf{m}$ convolves with the window vectors ($k$-grams) at each position. Each element $c_j$ of the feature map for window vector $w_j$ is produced as follows:

\begin{equation}
    c_j = f(\mathbf{m}w_j + b)
\end{equation}

where $b$ is a bias term and $f$ is the ReLU nonlinear transformation function. We use $n$ filters to generate $n$ feature maps. To be more specific, the output from convolution operation with windeow size $k$ and $n$ filters is: $C^{s1} \in R^{(L-k+1)\times n}$. We assume there are $v$ different sizes of filters, so the output is: $C^{s1}, C^{s2} \dots C^{sv}$. Each of the output is subsequently fed into the LSTM  component individually. As LSTM is specified for sequence input while pooling operators break the sequence organization due to the discontinuous selected features, we do not use these operators after the convolution operation.

\textbf{LSTM Component.} We adopt the LSTM model \cite{goodfellow2016deep} to learn the long-range dependencies from higher-order sequential features generated by the CNN component of our model. Formally, the transition functions are defined as follows:

\begin{equation}
\begin{aligned}
\label{eqn:bilstm}
&i_t = \sigma(W_i \cdot [h_{t-1},x_t] + b_i) \\
&f_t = \sigma(W_f \cdot [h_{t-1},x_t] + b_f) \\
&o_t = \sigma(W_o \cdot [h_{t-1},x_t] + b_o) \\
&c_t = f_t \odot c_{t-1} + i_t \odot tanh(W_c \cdot [h_{t-1},x_t] + b_c) \\
&h_t = o_t\odot tanh(c_t)
\end{aligned}
\end{equation}

where $x_t$ denotes the current input feature representation from $C^{si}$ and $x_t \in \mathbb{R}^n$; $h_{t-1}$ is the old hidden state at previous time $t-1$; $i_t$, $o_t$, $f_t$ represent the values of an input gate, an output gate, a forget gate at time step $t$, respectively. We use $z$ as the number of hidden states in the LSTM. These gates collectively decide how to update the current memory cell $c_t$ and the current hidden state $h_t$. $W_i, W_f ,W_o, W_c  \in \mathbb{R}^{z \times (n+z)}$ are the weighted matrices and $b_i, b_f , b_o, b_c \in \mathbb{R}^z$ are biases, which parameterizing the transformations. Here, $\sigma$ is the logistic sigmoid function that has an output in [0, 1], $tanh$ denotes the hyperbolic tangent function that has an output in [-1, 1], and $\odot$ denotes the element-wise multiplication. Attention mechanism is subsequently applied on the learned hidden states to capture the informative words that contribute more towards detecting hate speech in social media post.

\textbf{Attention Mechanism.} Not all words in a post contribute equally to the detection of hate speech. We adapt the word attention proposed by Yang et al. \cite{yang2016hierarchical} to emphasis the important word features extracted from LSTM that aided hate speech detection and aggregate the representation of these informative word features to form the post's latent representations.

Specifically, we let $H \in \mathbb{R}^{z\times l}$ be a matrix consisting of hidden vectors $[h1, \cdots, h_l ]$ that the C-LSTM model produced, where $z$ is the size of hidden layers and $l$ is the length of the output. The length $l$ can be represented as $L-k+1$, when the window size of filters is $k$. The final hidden state of LSTM, $h_l$, contains information of the whole sentence. To implement self attention over the sentence, we consider the relation between each word and the meaning of the whole sentence. As a consequence, we use the final hidden state to attend to hidden state of each word to obtain the importance of each word for detecting hate speech. The attention mechanism will produce an attention weight vector $\alpha$ and the post's latent representations $x^p$.
    \begin{equation}
        M = tanh(W_H H + W_h h_L + b_h) 
    \end{equation}
    \begin{equation}
        \alpha = softmax(w^T M)
    \end{equation}
    \begin{equation}
        x^p = H\alpha^T
    \end{equation}

where, $M \in \mathbb{R}^{a\times l}$ , $w \in \mathbb{R}^a$ are
projection parameters and $a$ is a middle dimension of computation. $\alpha \in \mathbb{R}^l$ is a vector consisting of
attention weights and $x^p$ is attended output where there is only single size of filters. As there are multiple filter sizes, we concatenate output from attention module to form the final latent representation of the post. As such, the post's semantic representations $x^p_{w\_g}$, $x^p_{w\_v}$, and $x^p_{w\_r}$, and sentiment representation $x^{p}_{s}$ are learned in similar fashion.

\begin{figure*}[t]
	\begin{center}
		\includegraphics[width=0.7\textwidth]{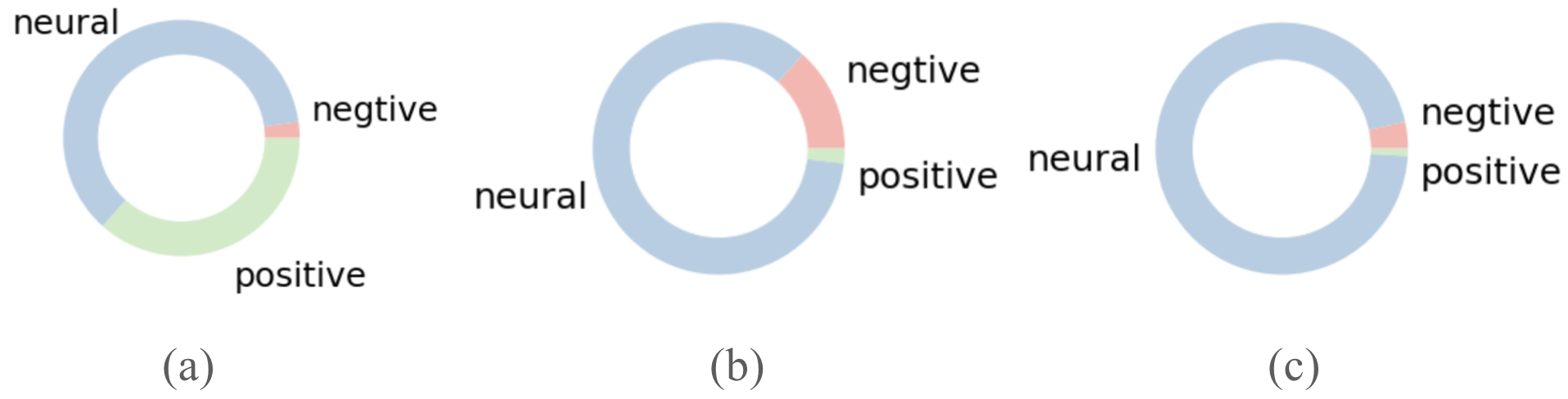}
	\end{center}
	\caption{Sentiment distribution of datasets. Figure a, b, c corresponds to the distribution of WZ-LS, DT and FOUNTA. Green in the pie chart denotes tweets labeled positive, orange denotes negtive and blue denotes neural.}
	\label{fig:distribution}
\end{figure*}

\subsection{Representation Fusion}
\label{sec:fusion}
After learning the latent semantic, sentiment, and topic representations, we perform operations to effectively combine the different post representations in the representation fusion layer.

In this layer, we adopt the Gate Attention Network to fuse information from multiple modalities. The Gate Attention Network proposed by Tang et al. \cite{DBLP:conf/aaai/TangCZ19} is effective for combining two vectors. Here we extend the gate fusion from two modalities to three modalities. The sentiment representation and topic representation are regarded as complementary information of semantic representation. So they are weighted by the interaction with semantic representation at first. Eventually, element-wise sum operation is performed to combine the  weighted sentiment and topic representation with the semantic representation:

    \begin{equation}
        att_w^s = \sigma(W_s (x_w^p + x_s^p))
    \end{equation}
    \begin{equation}
        att_w^t = \sigma(W_t (x_w^p + x_t^p))
    \end{equation}
    \begin{equation}
        x_J^p = x_w^p + att_w^s x_s^p + att_w^t x_t^p
    \end{equation}

where $W_s$ and $W_t$ are weighted matrices. The resulting joint representation $x_J^p$ is a high level representation of the post and can be used as features for hate speech multi-class classification in the softmax layer.



\subsection{Implementation Details}

We use the same embedding size $d = 300$ for pre-trained word embeddings and sentiment embedding. Additionally, we have added a dropout layer with $150\%$ dropout after the embedding layer for regularization and $20\%$ dropout for fully connected layer. Each CNN layer has three kinds of filter window size with 50 filters each. The number of hidden states in LSTM is set as 200. We use ADAM \cite{kingma2014adam} as the optimizer with a learning rate of $0.001$ to train our model.

\section{Dataset}
\label{sec:data}
We evaluate \textsf{DeepHate} on three publicly available datasets and one dataset reconstructed by merging tweets of the three datasets. The first three datasets, namely, WZ-LS \cite{ParkF17}, DT \cite{DavidsonWMW17}, and FOUNTA \cite{FountaDCLBSVSK18}, are widely used in hate speech detection studies. The merged dataset was constructed by concatenating the three datasets and removing the spam tweets. Table \ref{tab:dataset} shows the statistical summary of the these datasets. The table clearly shows that these datasets are diverse in both size and nature, thus allows us to evaluate our proposed model more comprehensively. 


\begin{table}[ht!]
  \caption{Statistic information about datasets in experiments}
  \label{tab:dataset}
  \begin{tabular}{c|c|p{14em}}
    \hline
    \textbf{Dataset} & \textbf{\#tweets} & \textbf{Classes (\#tweets)}\\\hline\hline
    WZ-LS & 13,202 & racism (82), sexism (3,332), both (21), neither (9,767)\\\hline
    DT & 24,783 & hate (1,430), offensive (19,190), neither (4,163)\\\hline
    FOUNTA & 89,990 & normal (53,011), abusive (19,232), spam (13,840), hate (3,907) \\\hline
    COMBINED & 114,120 & normal (66,941), inappropriate (47,194)\\\hline
\end{tabular}
\end{table}

{\bf WZ-LS dataset}. Park et al. \cite{ParkF17} combined two Twitter datasets \cite{waseem2016hateful,waseem2016you} to form the WZ-LS dataset. The dataset breaks down the types of hate speech into four classes: racism, sexism, both, and neither. As only the ids of the tweets are released in \cite{ParkF17}, we retrieve the text of the tweets using Twitter's APIs. However, some of the tweets have been deleted by Twitter due to their inappropriate content. Thus, our dataset is slightly smaller than the original dataset reported in \cite{ParkF17}.


{\bf DT dataset}. Davidson et al. \cite{DavidsonWMW17} argued that hate speech should be differentiated from offensive tweets; some tweets may contain hateful words but may be just offensive and did not meet the threshold of classifying them as hate speech. The researchers constructed the DT Twitter dataset, which manually labeled and categorized tweets into three categories: offensive, hate, and neither.


{\bf FOUNTA dataset}. The FOUNTA dataset is recently published in \cite{FountaDCLBSVSK18}. It's a human-annotated dataset that went through two rounds of annotations. In the first round, annotators are required to classify tweets into three categories: normal, spam, and inappropriate. Subsequently, the annotators were asked to refine further the labels of the tweets in the ``inappropriate'' category. Specifically, the final version of the dataset consists of four classes: normal, spam, hate, and abusive. We found that there were duplicated tweets in FOUNTA dataset as the dataset annotators have included retweets in their dataset. For our experiments, we remove the retweets resulting in the distribution in Table \ref{tab:dataset}.


{\bf COMBINED dataset}. The COMBINED dataset is an aggregation of all inappropriate tweets including offensive and hate tweets, and normal tweets in three datasets above. We postulate that the aggregated dataset is closer to the real-world application as the social media platform provider is likely to be interested in detecting and reducing both hate and offensive tweets. The COMBINED dataset is the largest dataset with diverse types of inappropriate tweets.


\section{Experiment}
\label{sec:experiment}
In this section, we will first describe the settings of experiments conducted to evaluate our \textsf{DeepHate} model. Next, we discuss the experiment results and evaluate how \textsf{DeepHate} fare against other state-of-the-art baselines. We conduct more in-depth ablation studies on the various components in the \textsf{DeepHate}. Empirical analyses on the sentiment, topics, and salient features that aided in hate speech detection are also conducted. Finally, we discuss some interesting case studies on how \textsf{DeepHate}'s strengths and limitations in hate speech detection. 


\subsection{Experiment Setup}
\textbf{Baselines.} For evaluation, we compare the \textbf{DeepHate} model with the following state-of-the-art baselines that utilized textual content for hate speech detection:

\begin{table}[t]
  \caption{Experiment results of DeepHate and baselines on WZ-LS dataset}
  \label{tab:wz-ls}
  \resizebox{0.9\columnwidth}{!}{
  \begin{tabular}{c|ccc}
    \hline
    \textbf{Model} & \textbf{Micro-Prec }& \textbf{Micro-Rec} &\textbf{ Micro-F1} \\\hline\hline
    CNN-W & 75.95 & 78.57 & 75.54 \\
    CNN-C & 54.77 & 74.01 & 62.95 \\
    CNN-B & 76.30 & 79.08 & 74.78 \\\hline
    LSTM-W & 75.39 & \textbf{79.52} & 74.52 \\
    LSTM-C & 74.82 & 78.13 & 71.95 \\
    LSTM-B & 54.77 & 74.01 & 62.95 \\\hline
    HybridCNN & 76.35 & 78.93 & 75.98 \\\hline
    CNN-GRU & 75.33 & 79.27 & 74.42  \\\hline
    DeepHate & \textbf{77.95} & 79.48 & \textbf{78.19}\\\hline
\end{tabular}}
\end{table}

\begin{table}[t]
  \caption{Experiment results of DeepHate and baselines on DT dataset}
  \label{tab:dt}
  \resizebox{0.9\columnwidth}{!}{
  \begin{tabular}{c|ccc}
    \hline
    \textbf{Model} & \textbf{Micro-Prec }& \textbf{Micro-Rec} &\textbf{ Micro-F1} \\\hline\hline
    CNN-W & 87.88 & 88.65 & 87.95 \\
    CNN-C & 60.53 & 77.43 & 67.60 \\
    CNN-B & 78.02 & 80.33 & 77.01 \\\hline
    LSTM-W & 88.08 & 89.08 & 87.87 \\
    LSTM-C & 77.21 & 79.88 & 76.47 \\
    LSTM-B & 59.97 & 77.44 & 67.60 \\\hline
    HybridCNN & 88.33 & 88.96 & 88.07 \\\hline
    CNN-GRU & 87.60 & 88.24 & 87.23 \\\hline
    DeepHate & \textbf{89.97} & \textbf{90.39} & \textbf{89.92} \\\hline
\end{tabular}}
\end{table}

\begin{table}[t]
  \caption{Experiment results of DeepHate and baselines on FOUNTA dataset.}
  \label{tab:founta}
  \resizebox{0.9\columnwidth}{!}{
  \begin{tabular}{c|ccc}
    \hline
    \textbf{Model} & \textbf{Micro-Prec }& \textbf{Micro-Rec} &\textbf{ Micro-F1}\\\hline\hline
    CNN-W & 78.26 & 79.71 & 78.27 \\
    CNN-C & 69.66 & 70.15 & 64.40 \\
    CNN-B & 52.01 & 58.41 & 50.64 \\\hline
    LSTM-W & 78.54 & 79.87 & 78.48 \\
    LSTM-C & 70.15 & 70.89 & 66.30 \\
    LSTM-B & 55.22 & 62.71 & 54.52 \\\hline
    HybridCNN & 78.34 & 79.24 & 77.73 \\\hline
    CNN-GRU & 78.62 & 80.17 & 78.39 \\\hline
    DeepHate & \textbf{78.95} & \textbf{80.43} & \textbf{79.09} \\\hline
\end{tabular}}
\end{table}

\begin{table}[t]
  \caption{Experiment results of DeepHate and baselines on COMBINED dataset.}
  \label{tab:total}
  \resizebox{0.9\columnwidth}{!}{
  \begin{tabular}{c|ccc}
    \hline
    \textbf{Model} & \textbf{Micro-Prec }& \textbf{Micro-Rec} &\textbf{ Micro-F1}\\\hline\hline
    CNN-W & 91.86 & 91.77 & 91.72 \\
    CNN-C & 79.56 & 79.06 & 78.50 \\
    CNN-B & 42.26 & 58.63 & 43.67 \\\hline
    LSTM-W & 91.54 & 91.54 & 91.52 \\
    LSTM-C & 82.46 & 80.64 & 79.70 \\
    LSTM-B & 64.93 & 65.50 & 63.85 \\\hline
    HybridCNN & 91.77 & 91.72 & 91.67 \\\hline
    CNN-GRU & 91.63 & 91.40 & 91.31 \\\hline
    DeepHate & \textbf{92.48} & \textbf{92.45} & \textbf{92.43} \\\hline
\end{tabular}}
\end{table}

\begin{itemize}
    \item \textbf{CNN}: Previous studies have utilized CNN to perform automatic hate speech detection and achieved good results \cite{badjatiya2017deep,gamback2017using,agrawal2018deep}. For baselines, we train three CNN models with different input embeddings: word embedding (CNN-W), character embedding (CNN-C), and character-bigram embedding (CNN-B).  
    \item \textbf{LSTM}: The LSTM model ,is another model that was commonly explored in previous hate speech detection studies \cite{badjatiya2017deep,agrawal2018deep,grondahl2018all}. Similarly, we train three LSTM models with different input embeddings: word embedding (LSTM-W), character embedding (LSTM-C), and character-bigram embedding (LSTM-B).
    \item \textbf{HybridCNN}: We replicate the HybridCNN model proposed by Park and Fung \cite{ParkF17} for comparison. The HybridCNN model trains CNN over both word and character embeddings for hate speech detection.
    \item \textbf{CNN-GRU}: The CNN-GRU model that was proposed in a recent study by Zhang et al. \cite{zhang2018detecting} is also replicated in our study as a baseline. The CNN-GRU model takes word embeddings as input.
\end{itemize}

\textbf{Sentiment Learning}. As mentioned in Section \ref{sec:sentiment}, we learn the post's sentiment using the VADER~\cite{HuttoG14} sentiment analysis tool. The learned sentiment will be used as labels (i.e, negative, positive, and neutral)  to train the sentiment representations of the post. Figure \ref{fig:distribution} shows the Sentiment distributions of the three datasets.

\textbf{Topic Modeling}. In learning post's topic representation, we set the number of topics in WZ-LS, DT, and FOUNTA datasets to 15, 10, 15, respectively.

\textbf{Training and Testing Set.} In our experiments, we adopt an 80-20 split, where for each dataset, 80\% of the posts are used for training with the remaining 20\% used for testing. 

\textbf{Evaluation Metrics.} Similar to most existing hate speech detection studies, we use micro averaging precision ({\bf Micro-Prec}), recall ({\bf Micro-Rec}), and F1 score ({\bf Micro-F1}) as the evaluation metrics. Micro averaging is preferred in our experiments as there are classes imbalance in the hate speech datasets. Also, five-fold cross-validation is used in our experiments, and the average results of the five folds are reported.






\subsection{Experiment Results}


Table \ref{tab:wz-ls} shows the experiment results on WZ-LS dataset while Tables \ref{tab:dt}, \ref{tab:founta} and \ref{tab:total} show that on DT, FOUNTA and COMBINED datasets respectively. In the tables, the highest figures are highlighted in \textbf{bold}. We observe that \textsf{DeepHate} outperformed the state-of-the-art baselines. Interestingly, we observed that the CNN and LSTM models with word embedding perform better than character-level embeddings input, suggesting that the semantics information in words is able to yield good performance for hate speech detection. The performance of character-level bi-gram embedding is worse among the three types of input feature embeddings. A possible reason may be that posts are short, and character-level bi-gram embedding may be ambiguous sometimes. We also noted that the basic CNN and LSTM with word embedding models are able to outperform some of the more complicated models such as HybridCNN and CNN-GRU in some occasions.  


It is worth noting that there are differences between the results of HybridCNN and CNN-GRU models in our experiments and the results that were reported in previous studies \cite{ParkF17,zhang2018detecting}. For instance, previous studies for HybridCNN \cite{ParkF17} and CNN-GRU \cite{zhang2018detecting} had conducted experiments on the WZ-LS dataset. However, we did not cite the previous scores directly as some of the tweets in WZ-LS has been deleted. Similarly, CNN-GRU was also previously applied to the DT dataset. However, in the previous work \cite{zhang2018detecting}, the researchers have cast the problem into binary classification by re-labeling the offensive tweets as non-hate. In our experiment, we perform the classification based on the original DT dataset \cite{DavidsonWMW17}. Therefore, we replicated the HybridCNN and CNN-GRU models and applied them to the updated WZ-LS dataset and original DT dataset.

\subsection{Ablation Study}

Our proposed DeepHate model is made up of several components. In this evaluation, we perform an ablation study to investigate the effects of different latent post representations. Specifically, we compare the following settings:

\begin{itemize}
    \item \textbf{Semantic}: We apply the \textsf{DeepHate} model using only the post's latent semantic representation $x^p_{w}$.
    \item \textbf{Topic+Semantic}: We apply the \textsf{DeepHate} model using only the post's latent semantic representation $x^p_{w}$ and topic representation $x^p_{t}$.
    \item \textbf{Sentiment+Semantic}: We apply the \textsf{DeepHate} model using only the post's latent semantic representation $x^p_{w}$ and sentiment representation $x^p_{s}$.
    \item \textbf{DeepHate}: Our original DeepHate model which utilized post's semantic, topic, and sentiment representations.
\end{itemize}

\begin{table}[t]
  \caption{Performance of various DeepHate components on WZ-LS dataset}
  \label{tab:ab-wz-ls}
  \resizebox{1\columnwidth}{!}{
  \begin{tabular}{c|ccc}
    \hline
    \textbf{Model} & \textbf{Micro-Prec }& \textbf{Micro-Rec} &\textbf{ Micro-F1}\\\hline\hline
    Semantic & 77.00 & 78.75 & 77.04 \\
    Topic+Semantic & \textbf{77.98} & 79.32 & 77.98\\
    Sentiment+Semantic & 77.09 & 78.62 & 77.35  \\\hline
    DeepHate & 77.95 & \textbf{79.48} & \textbf{78.19}\\\hline
\end{tabular}}
\end{table}

\begin{table}[t]
  \caption{Performance of various DeepHate components on DT dataset}
  \label{tab:ab-dt}
  \resizebox{1\columnwidth}{!}{
  \begin{tabular}{c|ccc}
    \hline
    \textbf{Model} & \textbf{Micro-Prec }& \textbf{Micro-Rec} &\textbf{ Micro-F1} \\\hline\hline
    Semantic & 89.44 & 90.24 & 89.49 \\
    Topic+Semantic & 89.56 & 90.28 & 89.64 \\
    Sentiment+Semantic & 89.59 & \textbf{90.39} & 89.64 \\\hline
    DeepHate & \textbf{89.97} & \textbf{90.39} & \textbf{89.92} \\\hline
\end{tabular}}
\end{table}

\begin{table}[t]
  \caption{Performance of various DeepHate components on FOUNTA dataset}
  \label{tab:ab-founta}
  \resizebox{1\columnwidth}{!}{
  \begin{tabular}{c|ccc}
    \hline
    \textbf{Model}& \textbf{Micro-Prec }& \textbf{Micro-Rec} &\textbf{ Micro-F1} \\\hline\hline
    Semantic & 78.68 & 80.40 & 78.57 \\
    Topic+Semantic & 78.77 & 80.45 & 78.62 \\
    Sentiment+Semantic & 78.88 & \textbf{80.53} & 78.79 \\\hline
    DeepHate & \textbf{78.95} & 80.43 & \textbf{79.09} \\\hline
\end{tabular}}
\end{table}

\begin{table}[t]
  \caption{Performance of various DeepHate components on COMBINED dataset}
  \label{tab:ab-total}
  \resizebox{1\columnwidth}{!}{
  \begin{tabular}{c|ccc}
    \hline
    \textbf{Model}& \textbf{Micro-Prec }& \textbf{Micro-Rec} &\textbf{ Micro-F1} \\\hline\hline
    Semantic & 92.26 & 92.23 & 92.20 \\
    Topic+Semantic & 92.33 & 92.30 & 92.27 \\
    Sentiment+Semantic & 92.32 & 92.28 & 92.25 \\\hline
    DeepHate & \textbf{92.48} & \textbf{92.45} & \textbf{92.43} \\\hline
\end{tabular}}
\end{table}

Table \ref{tab:ab-wz-ls}, \ref{tab:ab-dt}, \ref{tab:ab-founta} and \ref{tab:ab-total} shows the results of our ablation studies on WZ-LS, DT, FOUNTA and COMBINED datasets respectively. Our post's latent semantic representations, which combined the latent representations learned from three pre-trained word embeddings, outperforms the LSTM or CNN model with randomly initialized word embedding. We postulate that the combination of three pre-trained word embeddings allows more expressive semantic representations and ultimately leads to better hate speech detection. The addition of the post's latent topic and sentiment representations to its semantic representation is observed to improve performance, suggesting that topic and sentiment information are useful in hate speech detection. Finally, our \textsf{DeepHate} model that utilizes the post's semantic, topic, and sentiment representations constantly outperforms the other configurations in the four datasets, suggesting that multi-faceted textual representations can improve hate speech detection.

\subsection{Salient Feature Analysis}

\begin{figure}[t]
	\begin{center}
	\resizebox{1\columnwidth}{!}{
	\begin{tabular}{c|c}
	\hline 
    \textbf{Label} & \textbf{Example Post} \\\hline
	    Sexism &  \includegraphics[height=0.3cm]{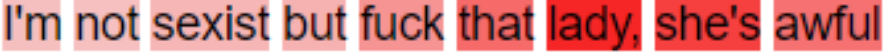}\\
	    Racism & \includegraphics[height=0.3cm]{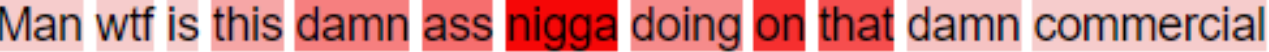}\\
	    Both   &  \includegraphics[height=0.3cm]{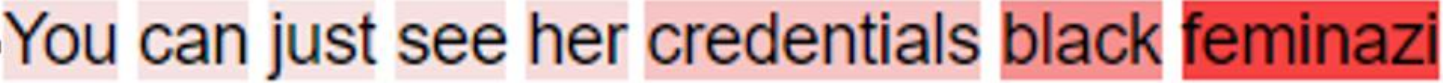}\\\hline
	\end{tabular}}
	\end{center}
	\caption{Visualized saliency scores on posts in WZ-LS dataset.}
	\label{fig:attention-vis-wz}
\end{figure}

\begin{figure}[t]
	\begin{center}
	\resizebox{0.8\columnwidth}{!}{
	\begin{tabular}{c|c}
	\hline 
    \textbf{Label} & \textbf{Example Post} \\\hline
	    Offensive   &  \includegraphics[height=0.26cm]{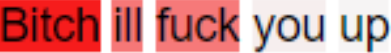}\\
	    Hate & \includegraphics[height=0.26cm]{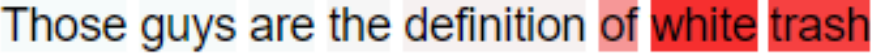}\\\hline
	\end{tabular}}
	\end{center}
	\caption{Visualized saliency scores on posts in DT dataset.}
	\label{fig:attention-vis-dt}
\end{figure}

\begin{figure}[t]
	\begin{center}
	\resizebox{1\columnwidth}{!}{
	\begin{tabular}{c|c}
	\hline 
    \textbf{Label} & \textbf{Example Post} \\\hline
	    Abusive &  \includegraphics[height=0.3cm]{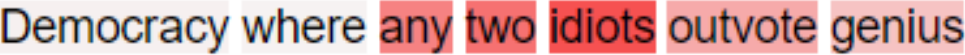}\\
	    Hate   &  \includegraphics[height=0.3cm]{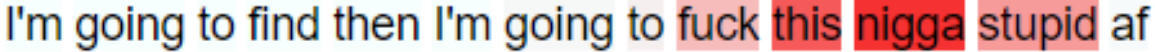}\\\hline
	\end{tabular}}
	\end{center}
	\caption{Visualized saliency scores on posts in FOUNTA dataset.}
	\label{fig:attention-vis-founta}
\end{figure}

To analyze how our \textsf{DeepHate} is able to extract important features from the textual information to perform hate speech detection, we examine the most salient sections of a single post's latent semantic representations. We adopt the saliency score defined by \cite{li2016visualizing} to measure the saliency of input features. The score indicates how sensitive a model is to the changes in the embedding input, i.e., in our case, how much a specific word, contributes to the final classification decision. Figure~\ref{fig:attention-vis-wz}, \ref{fig:attention-vis-dt}, and \ref{fig:attention-vis-founta} show the visualized computed saliency scores for WZ-LS, DT, and FOUNTA dataset respectively. Note that the more salient that section of the post's input feature representation, the darker is its shade of red. 

From the visualizations, we notice that the salient words highlighted in a post are descriptive of its true-label. For example, in Figure~\ref{fig:attention-vis-wz}, the words ``lady she's awful'' are highlighted for the post labeled as sexism. Similarly, words such as ``nigga'' are salient in the racism post. More interesting, words that are indicative of both sexism and racism are highlighted for posts labeled to contain both racism and sexism. Similar observations are made for other datasets. For instance, in the DT dataset, Figure~\ref{fig:attention-vis-dt} visualization shows that offensive lexicons are differentiated from hateful ones. The salient feature analysis demonstrates \textsf{DeepHate}'s capabilities to extracting critical textual features for hate speech detection. 

The extracted critical textual features also provide some form of explanation to \textsf{DeepHate}'s classification decision; it highlights the keywords which may suggest some form of hate or abusive language in the text. Besides achieving excellent detection performances, hate speech detection models should also be explainable to aid the content moderator in providing provide reasons for content removal. To the best of our knowledge, this is the first deep-learning-based hate speech detection study that supports explainability.

\subsection{Case Studies}
To gain a better understanding of the difference between \textsf{DeepHate} and baseline models, we qualitatively evaluate the models by examining some example posts and their classification results. Table \ref{tbl:casestudies-wz}, \ref{tbl:casestudies-dt}, and \ref{tbl:casestudies-founta} show example posts from WZ-LS, DT, and FOUNTA datasets respectively. In each example post, we display the true label and predicted labels from \textsf{DeepHate} and the best baseline, CNN-GRU. The correct predictions are marked in green font, while the incorrect predictions are represented in red font.

From the example posts, we observed that the posts' sentiments have a  profound impact on \textsf{DeepHate}'s hate speech detection performance. For instance, for posts with ``normal'' or ``non-hate'' true labels speech (e.g., post 1, 9, 15), \textsf{DeepHate} is correctly classified the posts while CNN-GRU made incorrect predictions. We postulate that sentiment information, i.e., neural sentiment, utilized by \textsf{DeepHate} has helped the model to minimize false negatives, i.e., labeling non-hate speech as hate speeches. However, the dependent on sentiment information also exposes a limitation of \textsf{Deephate} model: the inaccuracy of the sentiment information may have an adverse effect on \textsf{DeepHate}'s performance. For example, in post 5, 11, 12, 16, 17, the posts' sentiments are labeled as neural, and \textsf{DeepHate} has wrongly classified these posts. The sentiment labels learned using VADER model are also inaccurate as examining the posts' text clearly suggests that these are posts with negative sentiment. Thus, for future work, we will explore utilizing or designing better sentiment classification techniques to boost \textsf{DeepHate} performance in hate speech detection.    

\begin{table*}[t]
\caption{Example WZ-LS dataset's posts evaluated using \textsf{DeepHate} and best baseline (i.e., CNN-GRU)}
\label{tbl:casestudies-wz}
\begin{tabular}{|c|p{7cm}|c|c|c|}
\hline
\textbf{S/N} &\multicolumn{1}{|c|}{\textbf{Post}} & \multicolumn{1}{c|}{\textbf{True Label}} & \multicolumn{1}{c|}{\textbf{CNN-GRU}} & \multicolumn{1}{c|}{\textbf{DeepHate}}        \\ \hline\hline
1 & Kats face the only thing worse than her cooking. & Neither & {\color[HTML]{FE0000}Sexist} & {\color[HTML]{009901}Neither} (Sentiment: Neural) \\\hline
2 & I'm annoyed to have missed this live need some b*tchy. & Sexist & {\color[HTML]{FE0000}Neither} & {\color[HTML]{009901}Sexist} (Sentiment: Negative) \\ \hline
3 & Man w*f is this d*mn ass nig*a doing on that d*mn commercial. & Racist & {\color[HTML]{FE0000}Sexist} & {\color[HTML]{009901}Racist} (Sentiment: Negative) \\\hline
4 & Kat is f*cking c*nt. & Sexist & {\color[HTML]{009901}Sexist} & {\color[HTML]{FE0000}Neither} (Sentiment: Negative) \\ \hline
5 & It's not the only thing. She's done lot of, I'm guessing. & Sexist & {\color[HTML]{FE0000}Neither} & {\color[HTML]{FE0000}Neither} (Sentiment: Neural)\\ \hline
6 & Would slap the smirk off Kat's face. & Neither & {\color[HTML]{FE0000}Sexist} & {\color[HTML]{FE0000}Sexist} (Sentiment: Neural ) \\\hline
\end{tabular}
\end{table*}

\begin{table*}[t]
\caption{Example DT dataset's posts evaluated using \textsf{DeepHate} and best baseline (i.e., CNN-GRU)}
\label{tbl:casestudies-dt}
\begin{tabular}{|c|p{7cm}|c|c|c|}
\hline
\textbf{S/N} & \multicolumn{1}{|c|}{\textbf{Post}} & \multicolumn{1}{c|}{\textbf{True Label}} & \multicolumn{1}{c|}{\textbf{CNN-GRU}} & \multicolumn{1}{c|}{\textbf{DeepHate}}        \\ \hline\hline
7 & B*tch shut the f*ck up goddam your sl*t b*tch wh*re nig*a. & Hate & {\color[HTML]{FE0000}Offensive} & {\color[HTML]{009901}Hate} (Sentiment: Negative) \\\hline
8 & Stop looking p*rn fag. & Offensive & {\color[HTML]{FE0000}Non-hate} & {\color[HTML]{009901}Offensive} (Sentiment: Negative) \\\hline
9 & Okay, I'm going to say it once comb yer beards. & Non-hate & {\color[HTML]{FE0000}Offensive} & {\color[HTML]{009901}Non-hate} (Sentiment: Neural) \\\hline
10 & Why do people even talk about white privilege when the majority of food stamp recipients are white people. & Hate & {\color[HTML]{009901}Hate} & {\color[HTML]{FE0000}Non-hate} (Sentiment: Neural) \\ \hline
11 & And f*ck you too ya, little b*tch. You look like Mexican sucking c*ck in ur profile. & Hate & {\color[HTML]{FE0000}Offensive} & {\color[HTML]{FE0000}Offensive} (Sentiment: Neural)\\ \hline
12 & The nig is fun to watch. Got to admit Ebola boy can speak broken Spanish too, lol. & Hate & {\color[HTML]{FE0000}Offensive} & {\color[HTML]{FE0000}Offensive} (Sentiment: Neural) \\ \hline
\end{tabular}
\end{table*}

\begin{table*}[t]
\caption{Example FOUNTA dataset's posts evaluated using \textsf{DeepHate} and best baseline (i.e., CNN-GRU)}
\label{tbl:casestudies-founta}
\begin{tabular}{|c|p{7cm}|c|c|c|}
\hline
\textbf{S/N} & \multicolumn{1}{|c|}{\textbf{Post}} & \multicolumn{1}{c|}{\textbf{True Label}} & \multicolumn{1}{c|}{\textbf{CNN-GRU}} & \multicolumn{1}{c|}{\textbf{DeepHate}}        \\ \hline\hline
13 & Hate this b*tch with all the hate on the world. & Hate & {\color[HTML]{FE0000}Abusive} & {\color[HTML]{009901}Hate} (Sentiment: Negative) \\\hline
14 & Slow replies annoy the hell out of me. & Abusive & {\color[HTML]{FE0000}Normal} & {\color[HTML]{009901}Abusive} (Sentiment: Negative) \\ \hline
15 & We have the worst maternal mortality rate of any modern country and that will kill women. What is wrong with them? & Normal & {\color[HTML]{FE0000}Hate} & {\color[HTML]{009901}Normal} (Sentiment: Neural)  \\ \hline
16 & Don't like to hear women call their baby their little men. This is just another gender storm in cup. & Normal & {\color[HTML]{009901}Normal} & {\color[HTML]{FE0000}Abusive} (Sentiment: Neural)\\ \hline
17 & Looks like both the Chinese and the India news media have temporary amnesia. & Hate & {\color[HTML]{FE0000}Normal} & {\color[HTML]{FE0000}Abusive} (Sentiment: Neural)\\ \hline
18 & Netizens are so f*ckin annoying and dumb. & Hate & {\color[HTML]{FE0000}Abusive} & {\color[HTML]{FE0000}Abusive} (Sentiment: Negative) \\ \hline

\hline
\end{tabular}
\end{table*}

\section{Conclusion and Future Work}
\label{sec:conclusion}
In this paper, we proposed a novel deep learning framework known as \textsf{DeepHate}, which utilized multi-faceted text representations for automatic hate speech detection. We evaluated \textsf{Deephate} on three publicly available real-world datasets, and our extensive experiments have shown that \textsf{Deephate} outperformed the state-of-the-art baselines. We have also empirically analyzed the \textsf{Deephate} model and provided insights into the salient features that best aided in detecting hate speech in online social platforms.  Our salient feature analysis provided some form of explanation to \textsf{Deephate}'s hate classification decision. For future works, we will like to incorporate non-textual features into \textsf{Deephate} model and improve the posts' sentiment and topic representations with more advanced techniques.

\bibliographystyle{ACM-Reference-Format}
\balance
\bibliography{DeepHate}

\end{document}